\documentclass{article}

\usepackage{arxiv}  
\usepackage{etoolbox}
\usepackage[utf8]{inputenc}
\usepackage[T1]{fontenc}
\usepackage{hyperref}
\usepackage{url}
\usepackage{booktabs}
\usepackage{amsfonts}
\usepackage{nicefrac}
\usepackage{microtype}
\usepackage{graphicx}
\usepackage{natbib}
\usepackage{doi}
\usepackage{authblk}

\usepackage{amsmath, amsthm, amssymb}
\usepackage{csquotes}
\usepackage{subcaption}
\usepackage{multirow}
\usepackage{algorithm}
\usepackage{algorithmic}
\usepackage[table,xcdraw]{xcolor}
\usepackage{adjustbox}
\usepackage{newfloat}
\usepackage{listings}
\usepackage{paralist}

\title{SQAP-VLA: A Synergistic Quantization-Aware Pruning Framework for High-Performance Vision-Language-Action Models}

\author[1]{Hengyu Fang}
\author[1]{Yijiang Liu}
\author[1]{Yuan Du}
\author[2]{Huanrui Yang}
\author[1]{Li Du}

\affil[1]{School of Electronic Science and Engineering, Nanjing University \\
          \texttt{\{hengyufang, liuyijiang\}@smail.nju.edu.cn, \{yuandu, ldu\}@nju.edu.cn}}
\affil[2]{University of Arizona \\
          \texttt{huanruiyang@arizona.edu}}

\date{}

\usepackage{etoolbox} 

\makeatletter
\patchcmd{\maketitle}
  {A PREPRINT}
  {}
  {}{}
\makeatother

\renewcommand{\shorttitle}{SQAP-VLA} 

\hypersetup{
    pdftitle={SQAP-VLA: A Synergistic Quantization-Aware Pruning Framework for High-Performance Vision-Language-Action Models},
    pdfsubject={cs.AI, cs.RO, cs.CV},
    pdfauthor={Your Name Here},
    pdfkeywords={Vision-Language-Action Models, Model Compression, Quantization, Token Pruning, Robotics, Embodied AI},
    colorlinks=true,
    linkcolor=black,
    filecolor=black,
    urlcolor=black,
    citecolor=blue
}

\begin{document}
\maketitle
\begin{abstract}
Vision-Language-Action (VLA) models exhibit unprecedented capabilities for embodied intelligence. However, their extensive computational and memory costs hinder their practical deployment. Existing VLA compression and acceleration approaches conduct quantization or token pruning in an ad-hoc manner but fail to enable both for a holistic efficiency improvement due to an observed incompatibility. This work introduces SQAP-VLA, the first structured, training-free VLA inference acceleration framework that simultaneously enables state-of-the-art quantization and token pruning. We overcome the incompatibility by co-designing the quantization and token pruning pipeline, where we propose new quantization-aware token pruning criteria that work on an aggressively quantized model while improving the quantizer design to enhance pruning effectiveness. When applied to standard VLA models, SQAP-VLA yields significant gains in computational efficiency and inference speed while successfully preserving core model performance, achieving a $\times$1.93 speedup and up to a 4.5\% average success rate enhancement compared to the original model.
\end{abstract}
\noindent\textbf{Code:} The code is available at \url{https://github.com/ecdine/SQAP-VLA}.


\section{Introduction}

Vision-Language-Action (VLA) models~\citep{chi2024diffusionpolicyvisuomotorpolicy} constitute a major advancement in embodied intelligence, achieving exceptional performance on tasks that require integrated perception, language understanding, and real-world interaction. These models have catalyzed a broad spectrum of pioneering research in the field~\citep{kim2024openvlaopensourcevisionlanguageactionmodel,li2024cogactfoundationalvisionlanguageactionmodel,brohan2023rt2visionlanguageactionmodelstransfer,black2024pi0visionlanguageactionflowmodel}. However, their substantial computational and memory demands stand in stark contrast to the requirements for low-latency, energy-efficient deployment in edge devices for robotic applications. Bridging this gap necessitates effective model compression strategies. In particular, quantization~\citep{gholami2021surveyquantizationmethodsefficient, nagel2021whitepaperneuralnetwork} and token pruning~\citep{Wang_2021} have emerged as promising approaches for facilitating the efficient deployment of VLA models on resource-constrained hardware.

Quantization is recognized as an effective and generally applicable technique for model compression, especially at low precisions such as W4A4 (4-bit weights and 4-bit activations)~\citep{lin2025qservew4a8kv4quantizationcodesign, liu2025fbquantfeedbackquantizationlarge, li2025svdquantabsorbingoutlierslowrank}. In theory, W4A4 quantization can reduce the model size to one quarter and dramatically decrease computational costs compared to full-precision models~\citep{zhao2024atomlowbitquantizationefficient, shao2024omniquantomnidirectionallycalibratedquantization}.
Besides quantization, token pruning~\citep{liu2023revisitingtokenpruningobject} directly reduces the computational load and is hardware-friendly, leading to substantial improvements in inference speed~\citep{kuzmin2024pruningvsquantizationbetter}.
At first glance, quantization and token pruning appear to be naturally orthogonal and additive, suggesting that their integration should seamlessly yield highly efficient models~\citep{liang2021pruningquantizationdeepneural}. However, a straightforward combination of these methods leads to severe performance degradation. This arises from a fundamental coupling between the two techniques:
Token pruning leaves the model with limited information to work with, making the model more sensitive to quantization. In contrast, quantization profoundly alters the statistical distribution of features employed in token pruning, such as attention scores. As a result, pruning strategies developed for high-precision models become ineffective or even invalid when naively applied to quantized networks.
This intrinsic incompatibility has been largely overlooked in prior research and presents a major barrier to the deployment of compact and effective VLA models in real-world scenarios.

\begin{figure*}[th]
    \centering
    \includegraphics[width=1\linewidth]{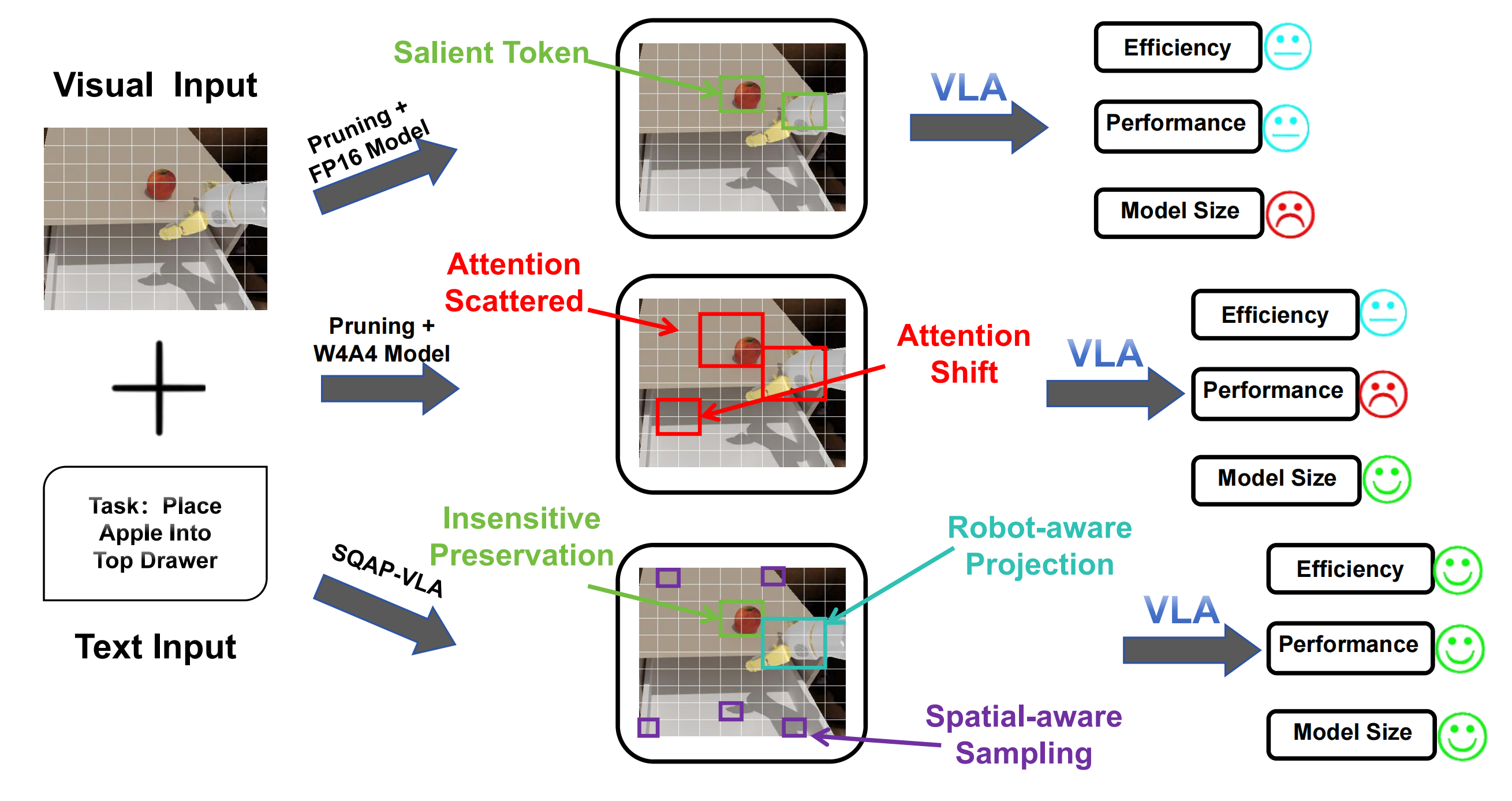}
    \caption{\textbf{Overview of SQAP-VLA framework.} SQAP-VLA resolves the incapability of token pruning on quantized VLA models via a quantization-aware pruning criteria. We propose insensitive preservation, robot-aware projection, and spatial-aware sampling to counter the scattered and shifted attention score of the quantized VLA model, enabling high performance, improved speed, and reduced model size with sparse tokens on a quantized VLA model.}
    \label{fig:enter-label}
\end{figure*}

To address the aforementioned incompatibility, we contend that an effective deployment of quantization and token pruning necessitates a principled co-design~\citep{Hawks_2021}, rather than a naive combination. We introduce a novel quantization-aware token pruning framework that jointly optimizes both compression techniques for robust performance~\citep{li2025spvlajointmodelscheduling}.
On the pruning side, we propose three strategies to ensure adaptation to quantization effects, as shown in Figure~\ref{fig:enter-label}. First, we identify and utilize quantization-insensitive pruning metrics by selectively retaining tokens with extreme attention scores~\citep{ye2024fitprunefasttrainingfree}, as their relative ordering is empirically robust to quantization noise and reliably preserves the most salient information. Second, we incorporate robot-aware prior protection~\citep{kleeberger2020singleshot6dobject}, i.e., leveraging the world coordinates of a robotic arm to ensure that its corresponding patch tokens are preserved, thereby safeguarding task-critical features. Third, we introduce spatially-aware sampling via Farthest Point Sampling~\citep{qi2017pointnetdeephierarchicalfeature} to maximize the spatial coverage of retained visual features and prevent information collapse. On the quantization side, we enhance the pruning-friendliness of the activation distributions by integrating Hadamard transforms~\citep{tseng2024quipbetterllmquantization}, which mitigate attention score distortion and yield more reliable pruning criteria. This synergistic framework outperforms conventional approaches, achieving superior performance retention under extreme compression and establishing a new standard for efficient, high-performance VLA model deployment on resource-constrained devices~\citep{zhu2024surveymodelcompressionlarge}.

Our primary contributions are:
\begin{itemize}
    \item We identify the intrinsic incompatibility between quantization and token pruning in Vision-Language-Action (VLA) models and propose a quantization-aware token pruning co-design framework that adapts token pruning strategies to quantization-induced feature distribution shifts, while also enhancing the effectiveness of pruning through quantization techniques.
    
    \item On the pruning side, we introduce quantization-insensitive preservation, robot-aware protection, and spatially-aware sampling, thereby adapting token pruning to quantized representations. On the quantization side, we propose the per-tensor Hadamard transformation to enhance attention distributions, thereby facilitating more reliable pruning criteria.
    
    \item Comprehensive experiments demonstrate that our integrated method not only resolves the failure of traditional pruning under extreme compression but also achieves the highest degree of performance, forging a viable pathway for the efficient, low-latency deployment of high-performance VLAs on resource-constrained devices.
\end{itemize}

\section{Related Work}

\subsection{Vision-Language-Action Models}
Vision-Language-Action (VLA) models~\citep{brohan2023rt1roboticstransformerrealworld, driess2023palmeembodiedmultimodallanguage, brohan2023rt2visionlanguageactionmodelstransfer} constitute a major advancement in embodied artificial intelligence.
These models enable end-to-end learning paradigms in which robots can interact with both visual environments and human language instructions.
Typical VLA models extend pretrained Vision-Language Models (VLMs)~\citep{liu2023visualinstructiontuning} by incorporating mechanisms for generating executable action sequences.
A prominent recent trend involves employing diffusion models~\citep{chi2024diffusionpolicyvisuomotorpolicy} as the action head, wherein the VLM processes visual inputs and textual commands while the diffusion process generates the corresponding action trajectories.
This paradigm is exemplified by models such as CogACT~\citep{li2024cogactfoundationalvisionlanguageactionmodel} and $\pi_0$~\citep{black2024pi0visionlanguageactionflowmodel}.
However, the large model size and the complexity of the decoding process present substantial challenges for deployment on resource-constrained platforms, highlighting the necessity for efficient solutions. In response, this work proposes a synergistic compression framework that integrates model quantization with token pruning, offering a promising direction for enabling efficient and high-performance VLA models.

\subsection{VLA Compression}
\paragraph{Quantization} is recognized as an effective and generally applicable technique for model compression.
QAIL~\citep{park2024quantizationawareimitationlearningresourceefficientrobotic} applies quantization-aware training to maintain robust VLA performance with 4-bit weights and activations (W4A4) yet necessitates costly retraining.
The large language model (LLM) typically constitutes the dominant component of VLA architectures.
Quantization techniques developed for LLMs~\citep{frantar2023gptqaccurateposttrainingquantization,lin2024awqactivationawareweightquantization,sun2025flatquantflatnessmattersllm,liu2025spinquantllmquantizationlearned} thus provide important references for VLA quantization.
For instance, QUIP\#~\citep{tseng2024quipbetterllmquantization}, which incorporates Hadamard transformations to mitigate activation outliers, is promising for VLA quantization. However, their group-wise quantization granularity poses limitations for efficient inference. In this work, we address this shortcoming by integrating the Hadamard transformation with tensor-wise quantization, thereby enhancing adaptability and efficiency in VLA scenarios.

\paragraph{Token Pruning} represents another widely adopted approach to model compression.
Previously in VLMs, mainstream methods have utilized attention scores to identify and prune insignificant tokens~\citep{chen2024imageworth12tokens, zhang2025textvisualattentionexploitingvisual, bolya2023tokenmergingvitfaster}.
Recent token pruning work in VLA models draws inspiration from these techniques. For example, SP-VLA~\citep{li2025spvlajointmodelscheduling} leverages the inherent redundancy in visual tokens. Meanwhile, EfficientVLA~\citep{yang2025efficientvlatrainingfreeaccelerationcompression} focuses on task relevance, compressing model input by identifying tokens most relevant to the current task while also incorporating a summary of historical information.
Cache-VLA~\citep{xu2025vlacacheefficientvisionlanguageactionmodel} takes another path by discovering and leveraging position-fixed tokens that are repeatedly used in sequential decision-making, thereby reducing redundant computation through token caching across time steps.
Mole-VLA~\citep{zhang2025molevladynamiclayerskippingvision} introduces a dynamic layer-skipping mechanism, which enables the model to bypass redundant layers based on input complexity.
Despite their successes in full-precision models, these methods do not consider the potential impact of quantization. We observe significant performance degradation when these techniques are naively combined with quantization. To systematically address this issue, we propose SQAP-VLA, a quantization-aware token pruning co-design framework.

\section{Methodology}
Our methodology introduces a novel quantization-aware token pruning strategy aimed at maintaining robust VLA performance in resource-constrained environments.
To counteract the criteria distortion caused by quantization, we propose three pruning strategies: quantization-insensitive preservation, robot-aware protection, and spatially-aware sampling. In parallel, we enhance the quantizer design to maximize pruning effectiveness via Hadamard transformation and tensor-wise quantization.

\subsection{Challenges of Token Pruning on Quantization}
\begin{figure}[tb]
    \centering
    \begin{subfigure}[b]{0.24\textwidth}
        \centering
        \includegraphics[width=\textwidth]{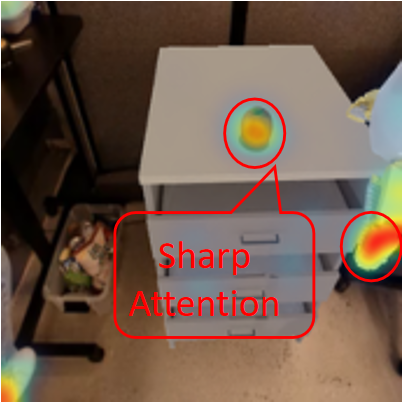}
        \caption{Full-precision model.}
        \label{fig:attention_before_quant}
    \end{subfigure}
    \hfill 
    \begin{subfigure}[b]{0.24\textwidth}
        \centering
        \includegraphics[width=\textwidth]{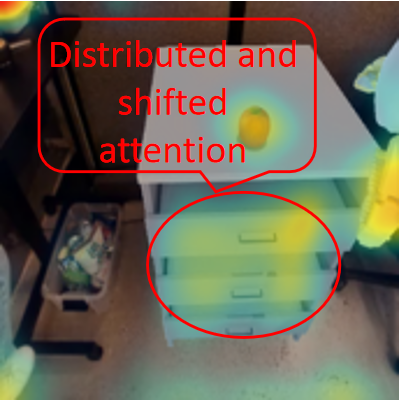}
        \caption{Quantized model.}
        \label{fig:attention_after_quant}
    \end{subfigure}
    \hfill
    \begin{subfigure}[b]{0.24\textwidth}
        \centering
        \includegraphics[width=\textwidth]{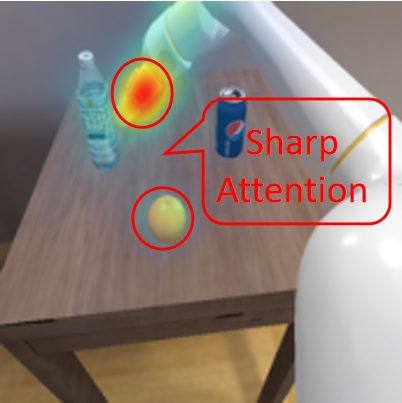}
        \caption{Full-precision model.}
        \label{fig:attention_before_quant2}
    \end{subfigure}
    \hfill
    \begin{subfigure}[b]{0.24\textwidth}
        \centering
        \includegraphics[width=\textwidth]{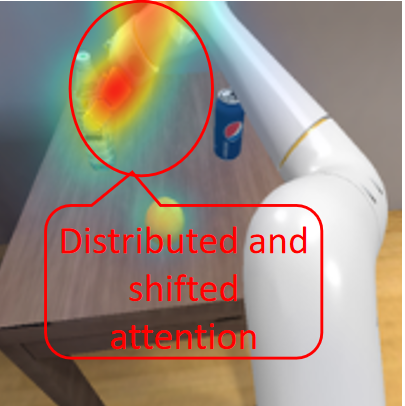}
        \caption{Quantized model.}
        \label{fig:attention_after_quant2}
    \end{subfigure}
    
    \caption{The attention heatmap before and after quantization. (a) and (c) Before quantization, attention is sharply focused. (b) and (d) After quantization, the attention becomes scattered and shifted.}
    \label{fig:attention_maps}
\end{figure}

Among existing token pruning approaches in VLMs, token importance is typically evaluated through the analysis of attention scores. In full-precision VLA models, task-relevant visual tokens generally attain high attention values.
As illustrated in Figure~\ref{fig:attention_before_quant}, the tokens corresponding to the ``apple'' and ``robot arm'' receive prominent attention scores for the task ``pick up the apple''.
However, if weights and activations of the VLA model are quantized to low precisions, noise will be introduced to the inference process, and attention scores will be distorted, resulting in both scattered (i.e., expanded focus areas) and shifted (i.e., focus on irrelevant regions) attention maps.
When token pruning methods rely on such severely degraded attention maps, their direct application produces suboptimal or even invalid performance.
This incompatibility between quantization and token pruning necessitates more meticulous designs to ensure robust and reliable model operation.

\subsection{Quantization-aware Pruning Strategies}
To address the aforementioned challenges, we propose three pruning strategies: quantization-insensitive token preservation, robot-aware token protection, and spatially-aware token sampling.
\paragraph{Strategy 1: Quantization-insensitive token preservation.}
While quantization perturbs the overall attention landscape, this distortion does not affect all tokens uniformly. Our empirical analysis reveals that the numerical shift primarily impacts tokens with mid-range attention scores, blurring the distinction between moderately and minimally important elements. Crucially, the indices of a small number of tokens with the highest-magnitude (``top-$k$") attention scores remain remarkably stable after quantization. These specific tokens are fundamentally important in VLA models, as they consistently correspond to the most task-critical visual elements like target objects or the robot's end-effector. Because the identity of these top-$k$ tokens is largely invariant to quantization noise, a `top-k` selection strategy provides a direct and robust mechanism to safeguard this vital information. Accordingly, we define the set of indices for these tokens, $\mathcal{K}_{\text{attn}}$, as:
\begin{equation}
\mathcal{K}_{\text{attn}} = \text{Top}_k(\mathbf{a}_q, k),
\end{equation}
where $\mathbf{a}_q \in \mathbb{R}^{N_v}$ is the attention weight vector from a task-query token to all $N_v$ visual tokens, and $\text{Top}_k(\cdot, k)$ is an operator that returns the indices of the $k$ largest values in the vector. With a $k$ small enough, the top-k selection is stable under quantization.

\paragraph{Strategy 2: Robot-aware token protection.}
To compensate for the reduced $k$ for quantization stability and create a truly quantization-invariant anchor, we leverage task-specific priors. We observed that attention scores in the original model consistently correlate with the visual tokens corresponding to the robotic arm. Instead of relying on degraded scores, we directly project the robot's known 3D world coordinates $(x_w, y_w, z_w)$ into 2D pixel coordinates $(u,v)$ using the camera's intrinsic ($\mathbf{K}$) and extrinsic ($[\mathbf{R}|\mathbf{t}]$) matrices:
\begin{equation}
    \lambda \begin{bmatrix} u \\ v \\ 1 \end{bmatrix} = \mathbf{K} [\mathbf{R} | \mathbf{t}] \begin{bmatrix} x_w \\ y_w \\ z_w \\ 1 \end{bmatrix}.
\end{equation}

These pixel coordinates are then mapped to discrete token coordinates $(t_u, t_v)$ based on the patch size $(P_w, P_h)$:
\begin{equation}
    t_u = \left\lfloor \frac{u}{P_w} \right\rfloor, \quad t_v = \left\lfloor \frac{v}{P_h} \right\rfloor.
\end{equation}

This allows us to form a ``protected" ring of tokens around a central token $\mathbf{t}_c$ (e.g., the robot's end-effector). The set of tokens in this ring, $\mathcal{K}_{\text{ring}}$, is selected using the Chebyshev distance ($\ell_{\infty}$ norm) within a radius $R_t$:
\begin{equation}
    \mathcal{K}_{\text{ring}} = \left\{ \mathbf{t} \in \mathcal{T} \;\middle|\; \|\mathbf{t} - \mathbf{t}_c\|_{\infty} \le R_t \right\}.
\end{equation}
This method, grounded in the robot's physical state, provides a stable and reliable mechanism for preserving the model's core visuomotor grounding, regardless of quantization errors.

\paragraph{Strategy 3: Spatially-aware token sampling.}
Analogous to the selective attention mechanism in human vision, we maintain high fidelity for critical points of interest while representing peripheral regions with reduced detail. After securing the high-importance tokens, we process the remaining tokens $\mathcal{T}_{\text{remain}} = \mathcal{T} \setminus (\mathcal{K}_{\text{attn}} \cup \mathcal{K}_{\text{ring}})$ to reduce redundancy. We apply Farthest Point Sampling (FPS) to select a spatially diverse subset of $m$ tokens from this remainder:
\begin{equation}
    \mathcal{K}_{\text{fps}} = \text{FPS}(\mathcal{T}_{\text{remain}}, m).
\end{equation}
This approach efficiently prunes redundant information while preserving broad spatial coverage, thereby leading to substantial acceleration in model inference. The final set of tokens retained is the union of these three sets: $\mathcal{K}_{\text{final}} = \mathcal{K}_{\text{attn}} \cup \mathcal{K}_{\text{ring}} \cup \mathcal{K}_{\text{fps}}$. In this allocation, $\mathcal{K}_{\text{ring}}$ represents a relatively fixed number of tokens, while the size of $\mathcal{K}_{\text{attn}}$ is the main component adjusted in proportion to the overall target pruning rate. The remaining token quota is then filled by $\mathcal{K}_{\text{fps}}$ to ensure the final distribution of unpruned tokens is spatially balanced.

\subsection{Pruning-Targeted Quantizer Enhancement}

\begin{figure}[htbp]
    \centering
    \begin{subfigure}[b]{0.45\textwidth}
        \centering
        \includegraphics[width=\textwidth]{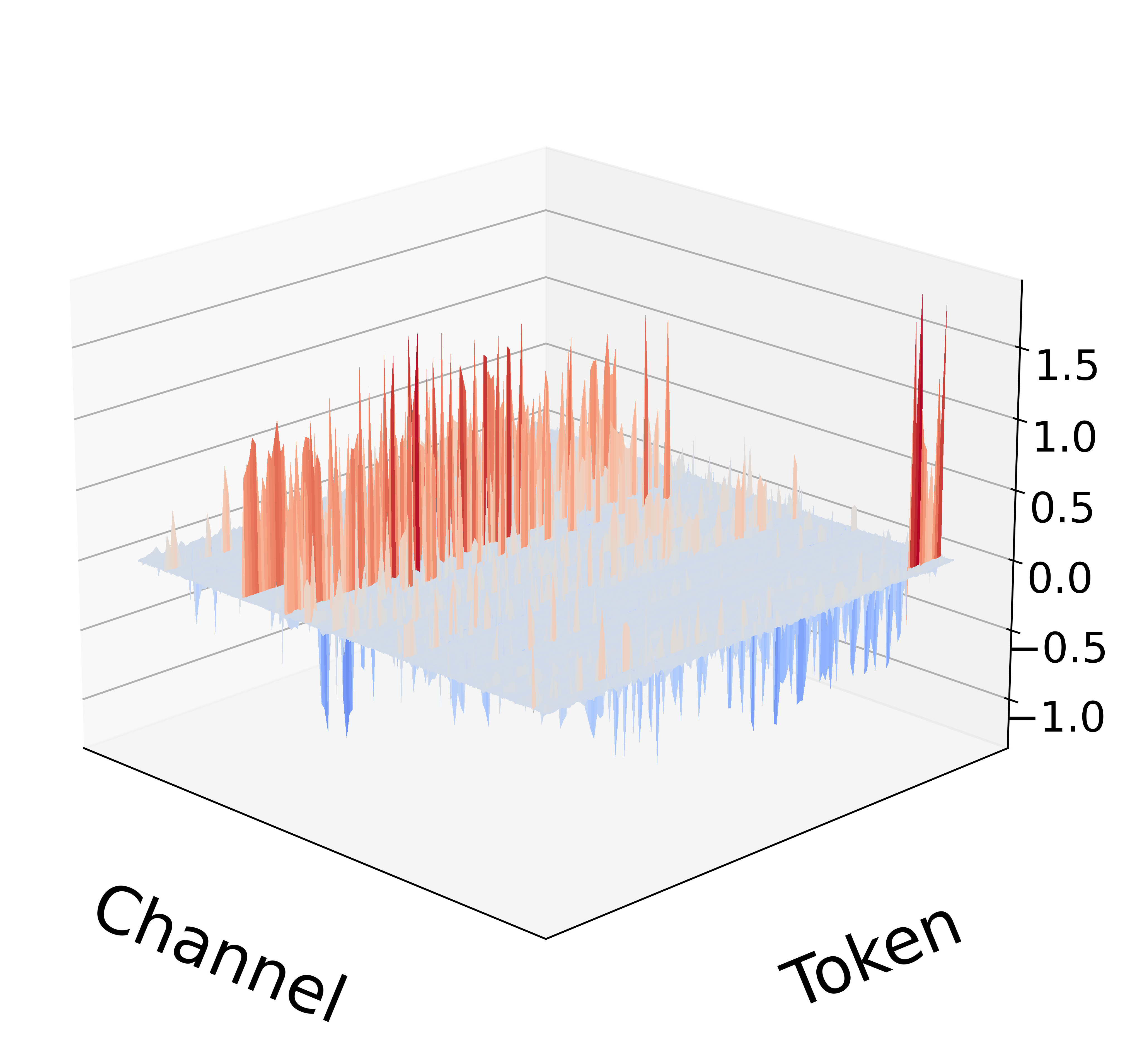}
        \caption{Original activation.}
        \label{fig:activations_before}
    \end{subfigure}
    \hfill
    \begin{subfigure}[b]{0.45\textwidth}
        \centering
        \includegraphics[width=\textwidth]{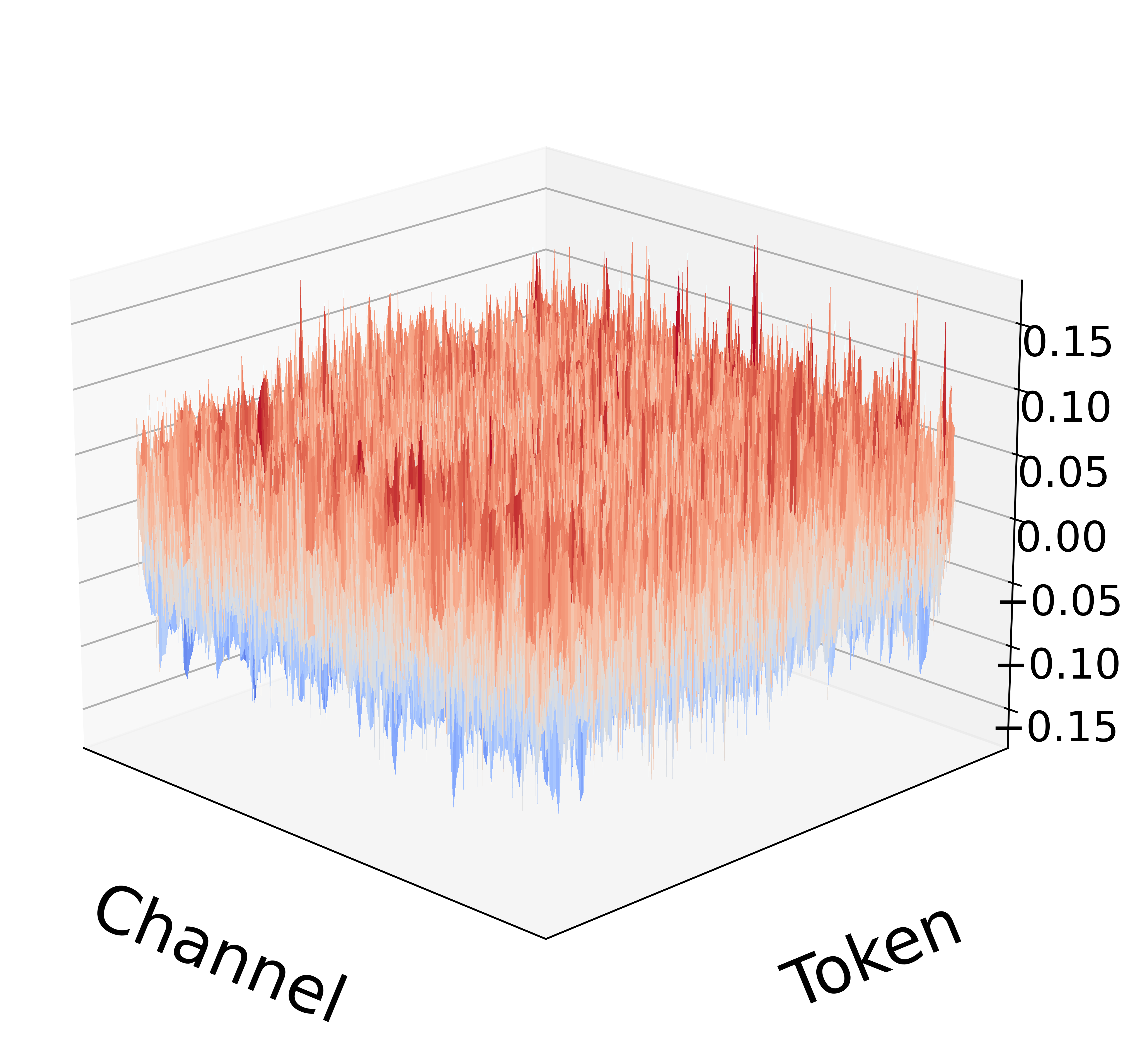}
        \caption{After Hadamard rotation.}
        \label{fig:activations_after}
    \end{subfigure}
    \caption{Visualization of activation distributions. (a) Original activations are dominated by large-magnitude outliers in specific channels. (b) After rotation, the energy from these outliers is uniformly distributed, eliminating extreme spikes.}
    \label{fig:hadamard_effect}
\end{figure}

The above pruning strategies partially alleviate the failure of token pruning under quantization.
However, the interpretability of the attention map, which enables effective token pruning, is still fundamentally disrupted by quantization-induced artifacts.
Specifically, quantization can severely distort the internal representations required for accurate token selection by degrading the quality of attention maps.
The attention mechanism depends on query and key vectors projected from input activations, i.e., $Q=W_q^TX$ and $K=W_k^TX$.
Our analysis reveals that these activations exhibit a highly skewed and asymmetric distribution. As shown in Figure~\ref{fig:activations_before}, a small set of fixed channels consistently produce values several magnitudes larger than the rest, introducing significant outliers into the activation landscape.
Traditional token-wise activation quantization is particularly susceptible to these outliers, resulting in considerable quantization errors and degraded attention representation.
To address this, we propose the use of channel-wise quantization for activations, which enables finer granularity and achieves improved quantization fidelity, especially for channels with smaller activation values.
However, the presence of large outlier channels continues to pose significant challenges for quantization. To further mitigate this issue, we take inspiration from LLM quantization QUIP\#~\citep{tseng2024quipbetterllmquantization} to use the Hadamard transformation on weights and activations of the query and key layers with the following formulation:
\begin{equation}
(W^TH^T)(HX) = W^T(H^T H)X = W^TX,
\end{equation}
where $H$ represents the Hadamard matrix.
The Hadamard transform effectively redistributes the activation energy more uniformly across all channels, thereby suppressing outlier effects~\citep{tseng2024quipbetterllmquantization}.
As visualized in Figure~\ref{fig:activations_after}, this transformation substantially smooths the activation landscape, resulting in improved reliability of the attention map and thereby enhancing token pruning performance.

\section{Experiments}

In this section, we present the experimental results and analysis. We begin by describing the experimental setup. Next, we demonstrate our main results with a highlight on efficiency. Finally, we conduct ablation studies on the pruning ratio and our proposed compression strategies.

\subsection{Experimental Setup}

\paragraph{Model Architecture}
Our experiments are centered on CogAct, a state-of-the-art Vision-Language-Action (VLA) model. Its architecture consists of a Prism-DinoSigLIP-224px~\citep{karamcheti2024prismaticvlmsinvestigatingdesign} vision encoder for perception, a Large Language Model (LLM) for high-level reasoning and instruction understanding, and a diffusion model for generating continuous action sequences. We evaluate our framework on two official variants of CogAct presented in the tables, Visual Matching and Variant Aggregation, to demonstrate the general applicability of our method.

\paragraph{Dataset and Training-Free Setup}
We leverage the official CogAct models, which come pre-trained on the large-scale Open X-Embodiment (OXE) dataset~\citep{embodimentcollaboration2025openxembodimentroboticlearning}. A key advantage of our framework, SQAP-VLA, is that it is entirely training-free. We apply our post-training quantization and pruning techniques directly to the publicly available model checkpoints, completely bypassing any need for costly retraining or fine-tuning.

\paragraph{Baselines}
We evaluate the effectiveness of our approach by comparing it with a comprehensive set of baselines. Specifically, we consider the original CogAct model in full precision (FP16), which serves as the primary reference for both performance and inference speed. Additionally, we benchmark our SQAP-VLA framework against several established token pruning methods applied to the FP16 model, including Random Dropping, FastV~\citep{chen2024imageworth12tokens}, VLA-Cache~\citep{xu2025vlacacheefficientvisionlanguageactionmodel}, and EfficientVLA~\citep{yang2025efficientvlatrainingfreeaccelerationcompression}. This comparison enables a rigorous assessment of our method relative to state-of-the-art acceleration techniques.

\paragraph{Evaluation Environment and Tasks}
All evaluations are conducted in a standard robotics simulation benchmark. We report performance on four challenging and representative manipulation tasks that require precise visuomotor control: Pick Coke Can, Move Near, Open/Close Drawer, and Place Apple in Top Drawer.

\paragraph{Evaluation Metrics}
To provide a comprehensive evaluation of our approach, we employ two primary metrics. The Success Rate (SR, expressed as a percentage) quantifies task performance by measuring the proportion of successful trials. Additionally, we report the speed-up, defined as the theoretical acceleration factor relative to the FP16 CogAct baseline. Together, these metrics capture both the effectiveness and efficiency of the evaluated models.

\paragraph{Implementation Details}
All our experiments, including quantization and pruning, are conducted in a training-free manner. The efficiency metrics (latency and memory) are benchmarked on a single NVIDIA RTX 3090 GPU. For our proposed method, SQAP-VLA, we apply our synergistic quantization-aware pruning to a W4A4 (4-bit weights and 4-bit activations) quantized model. Based on our ablation studies (Table~\ref{tab:ablation_pruning_avg}), we use a token pruning ratio of 0.4, which was found to yield the optimal trade-off between performance and efficiency.

\subsection{Main Results}

\begin{table*}[tb]
\centering
\setlength{\tabcolsep}{4.5pt} 
\caption{\textbf{Performance on the Visual Matching scenario.} We compare FP16 pruning methods against our method, which applies pruning to a W4A4 quantized model. Our approach achieves the best average performance. O/C Drawer refers to the Open/Close Drawer task, and BOPs stands for Basic OPerations. Our SQAP-VLA achieves the best average success rate and efficiency.}
\label{tab:main_results_vm}
\begin{tabular}{lccccccccr}
\toprule
\multirow{2}{*}{\textbf{Method}} & \multirow{2}{*}{\textbf{Quant.}} & \multicolumn{5}{c}{\textbf{Visual Matching Performance}} & \multirow{2}{*}{\textbf{Speed-up}} & \multirow{2}{*}{\textbf{BOPs}} \\
\cmidrule(lr){3-7}
& & \textbf{Pick Coke} & \textbf{Move Near} & \textbf{O/C Drawer} & \textbf{Place Apple} & \textbf{Average} & & \\
\midrule
CogACT (Baseline) & FP16 & 91.3 & 85.0 & 71.8 & 50.9 & 74.8 & 1.0$\times$ & 100.0\% \\
Random Dropping & FP16 & 9.7 & 20.4 & 53.5 & 0.0 & 20.9 & 1.2$\times$ & 58.5\% \\
FastV & FP16 & 92.6 & 81.4 & 69.8 & 52.4 & 74.1 & 1.21$\times$ & 42.0\% \\
VLA-Cache & FP16 & 92.0 & 83.3 & 70.5 & 51.6 & 74.4 & 1.38$\times$ & 80.1\% \\
EfficientVLA & FP16 & 95.3 & 83.3 & 70.3 & 56.5 & 76.4 & 1.59$\times$ & 45.1\% \\
\rowcolor[HTML]{F5D7D6} 
\textbf{SQAP-VLA} & \textbf{W4A4} & \textbf{94.7} & \textbf{85.5} & \textbf{72.2} & \textbf{64.8} & \textbf{79.3 } & 1.93$\times$ & 26.3\% \\
\bottomrule
\end{tabular}
\end{table*}

\begin{table*}[tb]
\centering
\setlength{\tabcolsep}{4.5pt} 
\caption{\textbf{Performance on the Variant Aggregation scenario.} Our method demonstrates superior or highly competitive performance against FP16 baselines, showcasing its effectiveness even after aggressive low-bit quantization.}
\label{tab:main_results_va}
\begin{tabular}{lccccccccr}
\toprule
\multirow{2}{*}{\textbf{Method}} & \multirow{2}{*}{\textbf{Quant.}} & \multicolumn{5}{c}{\textbf{Variant Aggregation Performance}} & \multirow{2}{*}{\textbf{Speed-up}} & \multirow{2}{*}{\textbf{BOPs}} \\
\cmidrule(lr){3-7}
& & \textbf{Pick Coke} & \textbf{Move Near} & \textbf{O/C Drawer} & \textbf{Place Apple} & \textbf{Average} & & \\
\midrule
CogACT (Baseline) & FP16 & 89.6 & 80.8 & 28.3 & 46.6 & 61.3 & 1.0$\times$ & 100.0\% \\
Random Dropping & FP16 & 4.0 & 16.1 & 15.6 & 0.0 & 8.9 & 1.20$\times$ & 58.5\% \\
FastV & FP16 & 91.4 & 78.6 & 27.6 & 50.6 & 62.1 & 1.19$\times$ & 42.0\%\\
VLA-Cache & FP16 & 91.7 & 79.3 & 32.5 & 45.8 & 62.3 & 1.37$\times$ & 82.6\% \\
EfficientVLA & FP16 & 94.8 & 77.6 & 28.4 & 51.9 & 63.2 & 1.57$\times$ & 45.1\% \\
\rowcolor[HTML]{F5D7D6} 
\textbf{SQAP-VLA} & \textbf{W4A4} & \textbf{92.8} & \textbf{80.6} & \textbf{27.2} & \textbf{57.0} & \textbf{64.4} & 1.93$\times$ & 26.3\% \\
\bottomrule
\end{tabular}
\end{table*}

The primary task performance results are summarized in Table~\ref{tab:main_results_vm} for the visual matching scenario and Table~\ref{tab:main_results_va} for the variant aggregation scenario.
The baseline for comparison is the pretrained CogAct model in full precision and without pruning.
The random dropping approach yields severely degraded performance, indicating its inadequacy for robust token pruning in VLA models.
Our SQAP-VLA consistently achieves state-of-the-art success rates and speedup ratios compared to alternative pruning methods, including FastV, VLA-Cache, and EfficientVLA. Remarkably, our model operates under W4A4 quantization, while the competing methods are evaluated in full-precision settings.
In the visual matching scenario, our proposed SQAP-VLA outperforms EfficientVLA by 2.9\% and exceeds the baseline by 4.5\%. In the variant aggregation scenario, our SQAP-VLA achieves a 3.1\% success rate improvement compared to the baseline.
In terms of computational efficiency, our approach delivers a $1.93\times$ speedup relative to the baseline and achieves a 36\% improvement over EfficientVLA.
We further observe that specialized token pruning techniques generally outperform the non-pruned baseline. For instance, both EfficientVLA and SQAP-VLA surpass the baseline in the visual matching scenario, and all the methods except random dropping achieve improved performance in the variant aggregation scenario.

\subsection{Efficiency Analysis}

\begin{figure}[tb]
    \centering
    \begin{subfigure}[b]{0.45\textwidth}
        \centering
        \includegraphics[width=\textwidth]{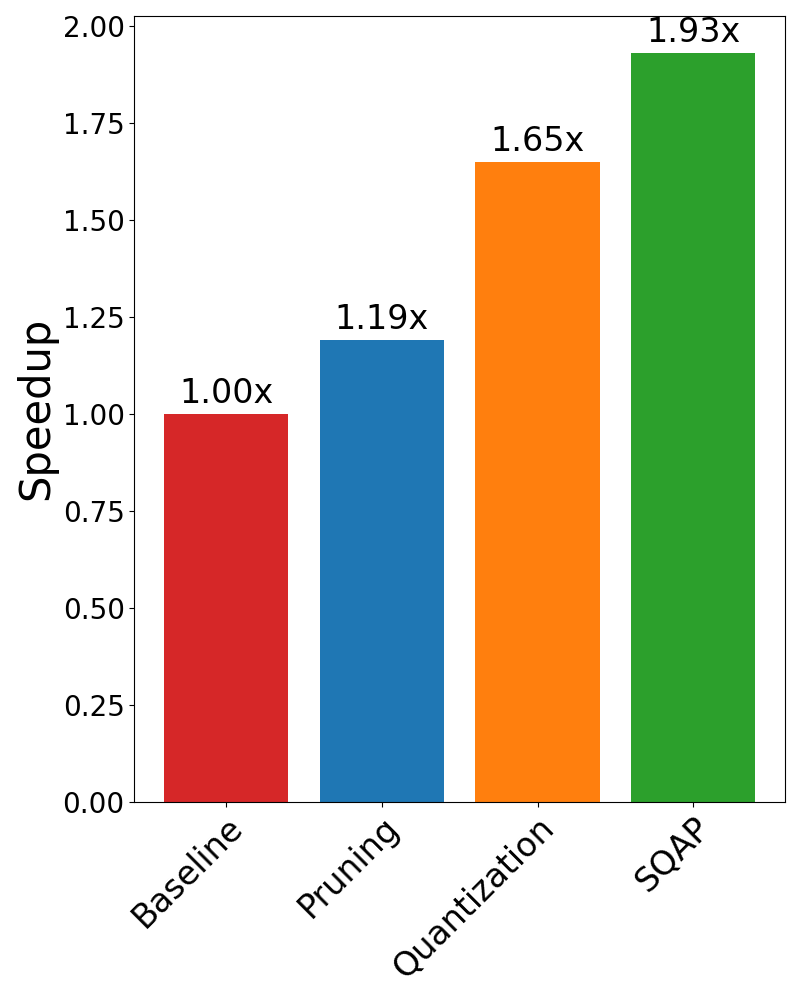}
        \caption{Speedup Analysis.}
        \label{fig:sqap_speedup}
    \end{subfigure}
    \hfill
    \begin{subfigure}[b]{0.45\textwidth}
        \centering
        \includegraphics[width=\textwidth]{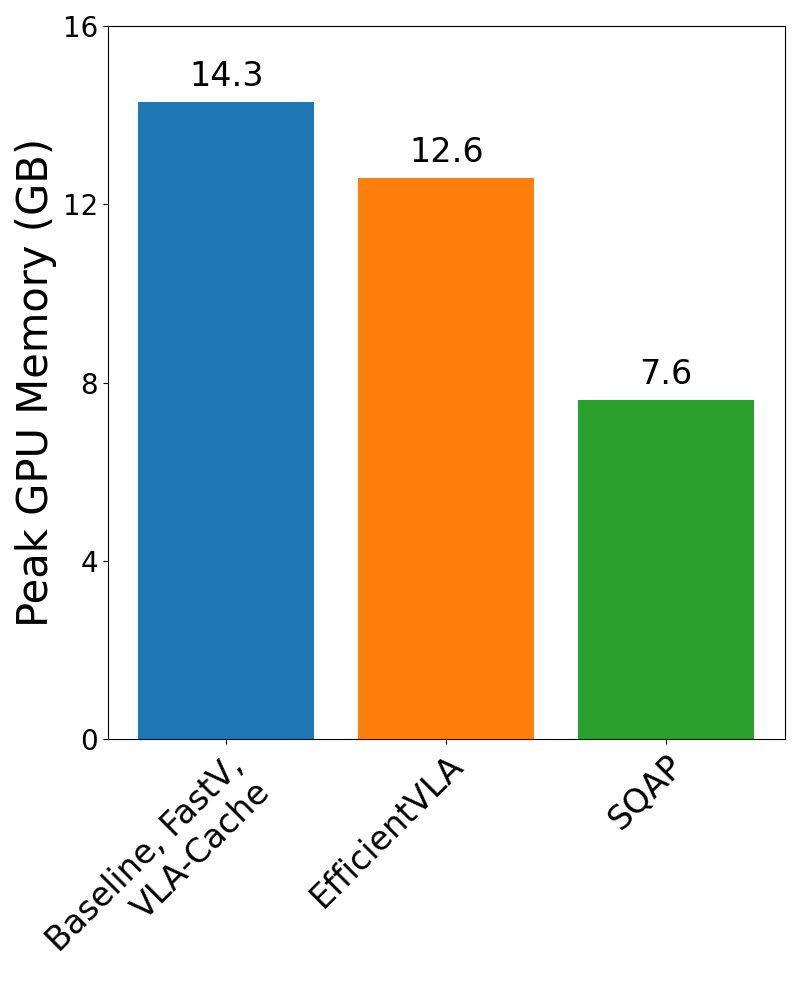}
        \caption{Peak GPU memory usage.}
        \label{fig:sqap_memory}
    \end{subfigure}
    \caption{\textbf{Latency and memory experiments.} (a) Ablation study on pruning and quantization of SQAP-VLA. (b) GPU memory comparison of the baseline, EfficientVLA, FastV, VLA-Cache, and SQAP-VLA.}
    \label{fig:sqap_performance}
\end{figure}

We conduct our efficiency analysis on a single NVIDIA RTX 3090 GPU using the Simpler simulator, focusing on the primary computational bottleneck in Vision-Language-Action (VLA) models: the Large Language Model (LLM) processing during the prefill stage.
As demonstrated in Figure~\ref{fig:sqap_speedup}, our SQAP-VLA method achieves a notable $1.93\times$ end-to-end system speedup over the FP16 baseline. This system-level gain is driven by a $2.56\times$ acceleration within the LLM backbone, which results from a synergistic combination of W4A4 quantization ($2.09\times$ speedup) and token pruning ($1.21\times$ speedup). In contrast to existing methods such as VLA-Cache and EfficientVLA, which focus exclusively on pruning to reduce the number of operations, our dual approach concurrently reduces both the input sequence length (number of operations) and the computational cost per operation.
This integrated strategy enables a more effective reduction of the total computational load, resulting in superior acceleration for large-scale parallel inference.
Beyond inference speedup, SQAP-VLA substantially mitigates GPU memory consumption. As illustrated in Figure~\ref{fig:sqap_memory}, the peak GPU memory usage decreases from 14.3 GB with the baseline model to 7.6 GB. This reduction in memory footprint is critical for resource-constrained edge devices, effectively overcoming the primary obstacle to on-device deployment of advanced VLA models.

\subsection{Ablation Study}

\begin{table}[ht]
    \centering
    \caption{\textbf{Ablation Study on Pruning Ratios}. We report the average success rate (\%) over all tasks. The best performance is highlighted in bold.}
    \label{tab:ablation_pruning_avg}
    \adjustbox{max width=\columnwidth}{
    \begin{tabular}{lcccccc}
    \toprule
    \multirow{2}{*}{\textbf{Model Variant}} & \multicolumn{5}{c}{\textbf{Pruning Ratio}} & \multirow{2}{*}{\textbf{Baseline (FP16)}} \\
    \cmidrule(lr){2-6}
    & \textbf{N/A (W4A4)} & \textbf{0.3} & \textbf{0.4} & \textbf{0.5} & \textbf{0.6} & \\
    \midrule
    Visual Matching & 71.78 & 76.55 & \textbf{79.30} & 78.43 & 76.85 & 74.80 \\
    Variant Aggregation & 58.18 & 61.35 & \textbf{64.40} & 63.63 & 61.18 & 61.30 \\
    \bottomrule
    \end{tabular}
    }
\end{table}

We conducted ablation studies on token pruning ratios varying from 0.3 to 0.6 and compression strategies including quantization, quantization-insensitive preservation, robot-aware protection, and spatially-aware sampling.

\paragraph{Pruning Ratios}
Table~\ref{tab:ablation_pruning_avg} presents the results of an ablation study evaluating various token pruning ratios.
The baseline model indicates the original model without quantization and pruning.
The “N/A” ratio corresponds to the W4A4 model without token pruning.
We investigate pruning ratios ranging from 0.3 to 0.6, where the value denotes the proportion of tokens pruned.
The W4A4 quantization harms the success rate by 3.02\% and 3.12\% respectively in the visual matching and variant aggregation scenarios.
Notably, models employing our pruning methods consistently outperform the original, unpruned model.
SQAP-VLA achieves the highest success rate at a pruning ratio of 0.4.
Thereby, we selected a ratio of 0.4 for our main method as it represents an excellent trade-off between high performance and the efficiency gains from more aggressive token pruning.

\paragraph{Compression Strategies}
Table~\ref{tab:ablation_strategy} presents an ablation study of our quantization and pruning strategies, designed to meticulously evaluate the contribution of each component. Our analysis commences with the uncompressed CogACT model as a baseline. As anticipated, applying a uniform W4A4 quantization results in a performance degradation, underscoring the inherent challenge of aggressive compression. While the subsequent integration of a generic quantization-insensitive token pruning affirms the basic compatibility of these techniques, it fails to fully recover the performance loss. The pivotal improvement stems from our context-aware strategies. The introduction of robot-aware token protection effectively reverses the performance decline. Ultimately, the addition of spatially-aware token sampling culminates in a final success rate of 79.30\% in visual matching scenarios, not only mitigating the initial quantization-induced deficit but decisively outperforming the full-precision baseline by a significant margin of 4.5\%. This empirically validates our core hypothesis that co-designing quantization with intelligent, task-centric pruning is crucial for developing highly efficient models that also achieve superior performance.

\begin{table*}[htbp]
\centering
\caption{\textbf{Ablation Study on Compression Strategies}. ``Baseline'' denotes the original full-precision model. We use 4-bit quantization on weights and activations. Three token pruning strategies are sequentially introduced to demonstrate the contribution of each component.}
\label{tab:ablation_strategy}
\adjustbox{max width=\textwidth}{
\begin{tabular}{clccccc}
\toprule
\multirow{2}{*}{\textbf{Compression Strategy}} & \multirow{2}{*}{\textbf{Model Variant}} & \multicolumn{4}{c}{\textbf{Success Rate (\%)}} & \multirow{2}{*}{\textbf{Average}} \\
\cmidrule(lr){3-6}
& & \textbf{Pick Coke Can} & \textbf{Move Near} & \textbf{Open/Close Drawer} & \textbf{Place Apple} & \\
\midrule
\multirow{2}{*}{Baseline} & Visual Matching & 91.3 & 85.0 & 71.8 & 50.9 & 74.80 \\
& Variant Aggregation & 89.6 & 80.8 & 28.3 & 46.6 & 61.30 \\
\midrule
\multirow{2}{*}{+ Quantization} & Visual Matching & 93.3 & 83.9 & 69.2 & 40.7 & 71.78 \\
& Variant Aggregation & 87.5 & 75.6 & 26.7 & 42.9 & 58.18 \\
\midrule
\multirow{2}{*}{+ Quant-Insensitive} & Visual Matching & 91.7 & 81.1 & 65.4 & 45.4 & 70.90 \\
& Variant Aggregation & 87.6 & 74.3 & 21.0 & 46.0 & 57.23 \\
\midrule
\multirow{2}{*}{\shortstack{+ Robot-Aware}} & Visual Matching & 92.0 & 85.0 & 69.2 & 49.7 & 73.98 \\
& Variant Aggregation & 90.6 & 75.6 & 26.7 & 48.7 & 60.40 \\
\midrule
\multirow{2}{*}{\shortstack{+ Spatially-Aware}} & Visual Matching & \textbf{94.7} & \textbf{85.5} & \textbf{72.2} & \textbf{64.8} & \textbf{79.30} \\
& Variant Aggregation & \textbf{92.8} & \textbf{80.6} & \textbf{27.2} & \textbf{57.0} & \textbf{64.40} \\
\bottomrule
\end{tabular}
}
\end{table*}

\section{Conclusion}
This paper tackles the deployment of large-scale Vision-Language-Action (VLA) models on resource-constrained platforms via joint token pruning and model quantization. We first identify a fundamental conflict between the two techniques: low-precision quantization catastrophically distorts the attention distributions that traditional token pruning methods rely upon, rendering them ineffective. To resolve this tension, we introduce a novel, training-free framework for the Synergistic Quantization-Aware Pruning of VLA models. Our approach is distinguished by a pruning strategy explicitly co-designed to be compatible with quantization that operates on quantization-insensitive signals to ensure effective token selections. Extensive experiments on the challenging ManiSkill2 benchmark validate the superiority of our method. It achieves a 1.93$\times$ speedup and reduces the GPU memory by over 73\% while maintaining or even surpassing the task performance of the full-precision baseline. This work presents a principled and effective solution that bridges the gap between high-performance VLA models and the stringent requirements of on-device deployment, offering a viable path toward more capable and responsive robotic systems. Future work could extend this synergistic framework to diverse architectures and explore dynamic, task-adaptive pruning policies.

\bibliographystyle{unsrtnat}
\bibliography{references}  

\end{document}